\documentclass[lettersize,journal]{IEEEtran}
\usepackage{amsmath,amsfonts}
\usepackage{algorithmic}
\usepackage{algorithm}
\usepackage[backref]{hyperref}
\usepackage{array}
\usepackage[caption=false,font=normalsize,labelfont=sf,textfont=sf]{subfig}
\usepackage{textcomp}
\usepackage{stfloats}
\usepackage{url}
\usepackage{lipsum}
\usepackage{verbatim}
\usepackage{graphicx}
\usepackage{cite}
\usepackage{caption}
\usepackage{multirow}
\usepackage{tabularx}
\usepackage{wrapfig}
\usepackage{booktabs}
\usepackage{tabularray}
\usepackage{wrapfig}
\usepackage{caption}
\usepackage{ulem} 
\usepackage{tabularx}
\usepackage{booktabs}
\usepackage{tabularray}  
\UseTblrLibrary{amsmath} 
\usepackage[table,xcdraw]{xcolor}
\usepackage{CJKutf8}
\begin{document}

\title{A Global Context Mechanism for Sequence Labeling}
\author{Conglei Xu, Kun Shen, Hongguang Sun, Yang Xu
\thanks{ Conglei Xu, Department of Computer Science, Aalborg University, Aalborg East, 9220, Denmark(email: cxu@cs.aau.dk)} 
\thanks{Kun Shen, Department of Electronic information Engineering,Rizhao Polytechnic, Rizhao, 276800, China(email: shenkun5410@rzpt.edu.cn)}
\thanks{Hongguang Sun, School of Information Science and Technology, Northeast Normal University, Changchun, 130117, China(email: sunhg889@nenu.edu.cn)
}
\thanks{Yang Xu, Customer Service Research and Development Division, China Unicom Software Research Institute, Beijing, 102676, China(email: xuy575@chinaunicom.cn)}
}

\markboth{Journal of \LaTeX\ Class Files,~Vol.~14, No.~8, August~2021}%
{Shell \MakeLowercase{\textit{et al.}}: A Sample Article Using IEEEtran.cls for IEEE Journals}

\maketitle

\begin{abstract}
Global sentence information is crucial for sequence labeling tasks, where each word in a sentence must be assigned a label. While BiLSTM models are widely used, they often fail to capture sufficient global context for inner words. Previous work has proposed various RNN variants to integrate global sentence information into word representations. However, these approaches suffer from three key limitations: (1) they are slower in both inference and training compared to the original BiLSTM, (2) they cannot effectively supplement global information for transformer-based models, and (3) the high time cost associated with reimplementing and integrating these customized RNNs into existing architectures. In this study, we introduce a simple yet effective mechanism that addresses these limitations. Our approach efficiently supplements global sentence information for both BiLSTM and transformer-based models, with minimal degradation in inference and training speed, and is easily pluggable into current architectures. We demonstrate significant improvements in F1 scores across seven popular benchmarks, including Named Entity Recognition (NER) tasks such as Conll2003, Wnut2017 , and the Chinese named-entity recognition task Weibo, as well as End-to-End Aspect-Based Sentiment Analysis (E2E-ABSA) benchmarks such as Laptop14, Restaurant14, Restaurant15, and Restaurant16. With out any extra strategy, we achieve third highest score on weibo NER benchmark. Compared to CRF, one of the most popular frameworks for sequence labeling, our mechanism achieves competitive F1 scores while offering superior inference and training speed. Code is available at: \url{https://github.com/conglei2XU/Global-Context-Mechanism}
 
\end{abstract}


\section{Introduction}
\IEEEPARstart{S}{equence} labeling tasks are fundamental to  information extraction, covering key applications such as named-entity recognition (NER), and Aspect-Based sentiment analysis(E2E-ABSA) tasks. These tasks play a critical role in downstream applications including knowledge graph construction \cite{xu2017}, improving information retrieval systems \cite{banerjee2019}, question-answering systems \cite{molla2006}, and more fine-grained sentiment analysis that targets specific aspects within text \cite{yan2021}.Unlike traditional classification tasks, such as sentence-level sentiment analysis that rely on the final sentence representation to make prediction, sequence labeling tasks demands precise token-level representations to accurately predict the label of each individual word. 

As natural language processing has entered the era dominated by large language models (LLMs), these models have demonstrated remarkable ability across many tasks \cite{sun2023,zhang2024, zhang2025}. However, when it comes to sequence labeling, there is a performance gap between supervised methods and large language models \cite{wang2023, xie2023}. These supervised methods generally combine pretrained transformers-which provide rich, contextualized word embeddings-with sequence modeling such as recurrent neural network (RNN) or conditional random file (CRF) to capture dependencies across the token sequence. For example, the hierarchical contextualized representation model \cite{luo2020} leverages BERT and BiLSTM alongside document-level information to set the state-of-the-art(SOTA) performance on the CoNLL-2003 benchmark\cite{tjong2003}. Similarly, Jana et al. \cite{strakova2019} utilized contextual word representation from ELMO, BERT, flair with BiLSTM and CRF, achieving new SOTA results on CoNLL-2003 benchmark. Furthermore, Li et al. \cite{li2019a} demonstrates the efficacy of combining BERT with BiLSTM to jointly extract aspect terms, categories, and sentiments, demonstrating the efficacy of this hybrid approach on sequence labeling.

Despite the widespread adoption and effectiveness of BiLSTM, it is well-recognized that BiLSTM representations of inner tokes lack sufficient global sentence context. This limitation arises from concating part of forward information and backward information as the representation for inner token, which inherently lack of global sentence information existed in the final backward and forward step. To mitigate this, several custom recurrent architectures have been proposed as an alternative for BiLSTM. Liu et al.\cite{Liu2019} and Meng et al.\cite{Meng2019}, for instance, introduced deep recurrent neural network transitional architectures to deepen the state transition path and assign a global sentence representation for each token. Zhang et al. \cite{Zhang2018}(S-LSTM) proposed the sentence-state LSTM (S-LSTM), which parallelizes the computation of local representations and global sentence representation simultaneously. Xu et al. \cite{xu2021} introduced Synergized-LSTM (Syn-LSTM), which combine the contextual and structural information derived from graph neural networks (GNN). Nevertheless, these sophisticated models come with practical drawbacks. These methods suffer from slower inference and training speeds compared to origin BiLSTM, which has been highly optimized in framework such as Pytorch and Tensforflow. Additionally, their logical and procedural complexity makes them time consuming to re-implement and integrate into existing framework. More importantly, while these custom RNN variants enhance global context integration at the recurrent modeling stage, they do not directly address the integration of global sentence information into pretrained transformer-based contextual embeddings. can be an alternative for BiLSTM for sequence labeling, they cannot inject the global sentence information into contextual representation generated from pretrained transformers. Li\cite{Li2020}argued that leveraging the self-attention mechanism can remedy lack of global sentence information in BiLSTM's inner word representations, however, this approach may inadvertently noise to representation of each token, since the interaction between a given word and all other words it not always semantically relevant, potentially contaminating the original embeddings.

\begin{figure*}[]
\centering
\captionsetup{justification=centering}

\includegraphics [width= 0.6\textwidth, height=10cm] {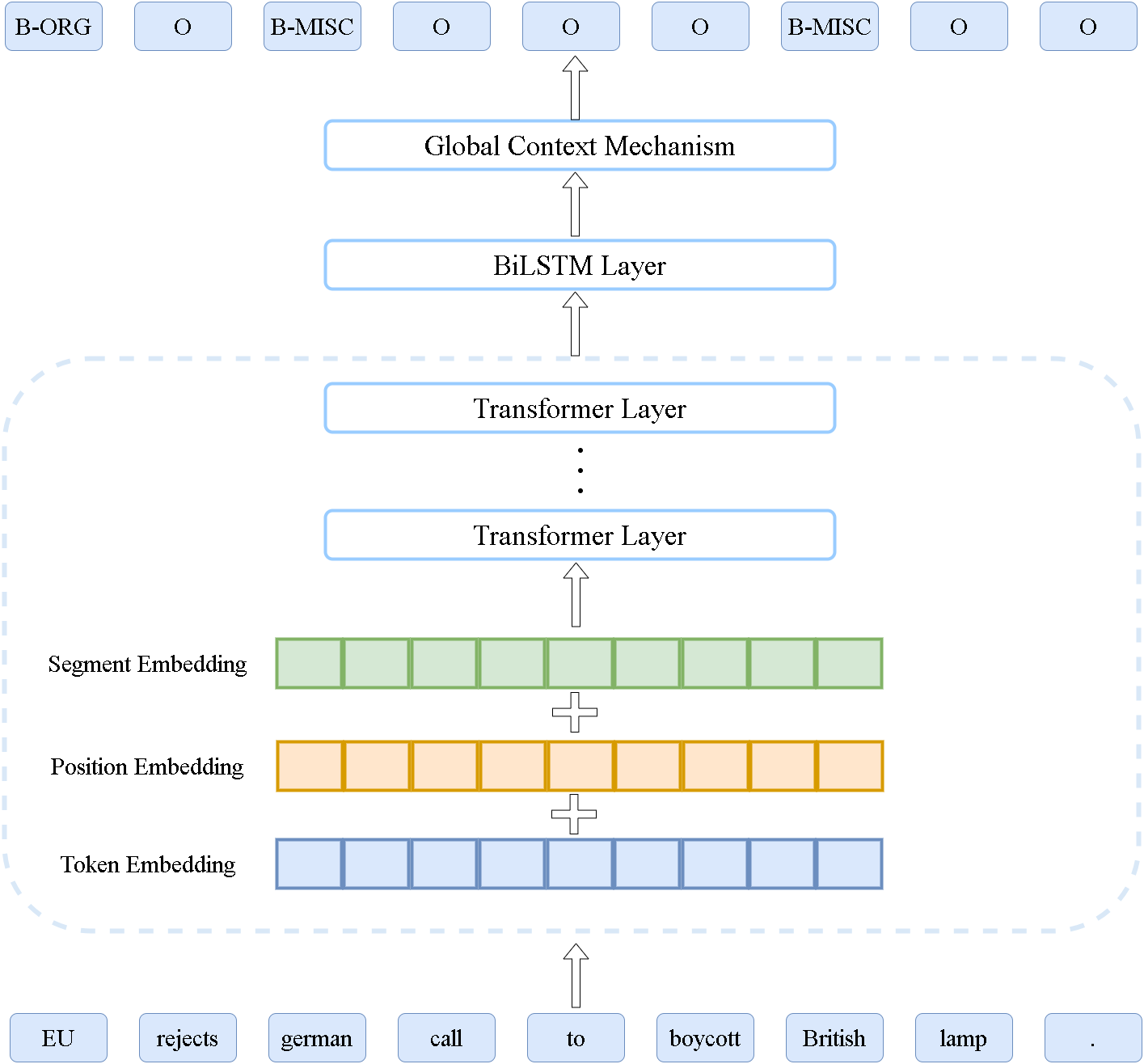}
\caption{Overview of the model architecture}
\label{fig_1}
\end{figure*}

To address these challenges, we design an efficient and general global context mechanism to enhance word representations with global sentence information for both BiLSTM and pretrained transformers. Our mechanism introduces only add two additional linear transformations, as illustrated in Figure1\ref{fig_2}. In particular, we use the representation of last forward step and forward step, or the [SEP] and [CLS] in pretrained transformers, as the forward and backward global sentence representations, respectively. These global representations are injected into the local representation of each token through element-wise weighting generated by a gate mechanism. Incorporating global sentence information in this way is beneficial for predicting each token's label, as it proved the model with comprehensive contextual cues that help disambiguate ambiguous cases in sequence labeling\cite{Li2020}. 

We evaluate our proposed mechanism on seven benchmarks covering two primary sequence labeling tasks, named entity recognition (NER) and end-to-end aspect-based sentiment analysis (E2E-ABSA). The datasets include CoNLL-2003, WNUT-2017, Weibo NER, Restaurant 14, Restaurant 15, Restaurant 16 and Laptop 14. Extensive experiments on these seven benchmarks upon different typical pretrained transformers-including BERT, Roberta, MacBET-suggest that the global context mechanism improve F1 score both for pretrained transformers and BiLSTM. Without employing any additonal training strategies, our methods achieves the third highes score on the Weibe benchmark. 

The main contributions of the paper are summarized as follows:

1. We proposed a general and efficient global context mechanism to enhance word representation for BiLSTM and pretrained transformers architectures.

2. Our mechanism improve F1 score for both BiLSTM and pretrained transformers with minimal degradation in training and inference speed and is easily pluggable into any architecture utilizing these backbone models. 

3. We implement a flexible sequence labeling framework that supports various pretrained transformers combined with BiLSTM and CRF.

4. We conduct thorough investigations on model complexity, the relationship between forward, backward global information with corresponding local information.

\section{Related Work}

\subsection{Tasks}

\textbf{End-to-End Aspect-based Sentiment Analysis (E2E-ABSA)}. Aspect-based sentiment analysis(ABSA) aims to identify the sentiment or opinion expressed by a user towards specific aspects \cite{mitchell2013,zhang2015} in user-generated text. The most widely used ABSA benchmark datasets originate from the SemEval datasets(SemEval-2014 task 4:
Aspect based sentiment analysis, SemEval-2015 task 12: Aspect based sentiment analysis,  SemEval-2016 task 5: Aspect based sentiment analysis\cite{pontiki2014,pontiki-etal-2015-semeval,pontiki2016}), where a few thousand review sentences with gold standard aspect sentiment are provided. As shown in Table\ref{tab:absa_settings}, end-to-end aspect-based sentiment analysis is a more challenging task in which aspect terms, aspect categories, and corresponding sentiments are jointly detected\cite{ma2017,schmitt2018,li2019b}.

\textbf{Named entity Recognition (NER).} NER is a basic task in information extraction, focusing on identifying and classifying named entities in text.  Current NER methods can be broadly categorized into following groups tagging-based \cite{ma2016}, span-based(Structured Prediction as Translation between Augmented Natural Languages) and generative-based models(Prompt Locating and Typing for Named Entity Recognition). In this work, we focus on tagging-based methods, which predict a label for each word. Give a sequence of of $ X={x_1, x_2, ..., x_n} $ with $ n $ tokens and its corresponding labels $ Y={y_1, y_2, ..., y_n} $ with the equal length, NER tasks aims to learn a parameterized function $ f_\theta : X -> Y $ from input tokens to task-special labels. 

\begin{table}[t]
\centering
\caption{ABSA and E2E-ABSA definition. Gold standard aspects and opinions are wrapped in \texttt{[]} and \texttt{<>} respectively. The subscripts N and P refer to aspect sentiment. \underline{*} or \uwave{*} indicates the association between the aspect and the opinion.}
\label{tab:absa_settings}
\begin{tabular}{lcc}
\toprule
\multicolumn{3}{l}{Sentence: \texttt{<\underline{Great}> [\underline{food}]\textsubscript{P} but the}} \\
\multicolumn{3}{l}{\texttt{[\uwave{service}]\textsubscript{N} is <\uwave{dreadful}>.}} \\
\midrule
Settings & Input & Output \\
\midrule
1. ABSA & sentence, aspect & aspect sentiment \\
2. E2E-ABSA & sentence & aspect, aspect sentiment \\
\bottomrule
\end{tabular}
\end{table}

\subsection{Neural Networks for Sequence Labeling}

\textbf{Pretrained transformers.} Transformers \cite{Vaswani2017} enable highly parallelized computation of sentence semantics using self-attention mechanisms. The BERT model \cite{Devlin2018} stacks transformer layers and has achieved state-of-art performance on a variety of natural language understanding tasks such as MultiNLI, GLUE, and SQuAD.Subsequent works have proposed optimizations of BERT, including RoBERTa \cite{liu2019b}, which refines hyperparameters and utilizes additional training data, and MacBERT \cite{Cui2020}, which modifies the masked language modeling objective to a language correction task.  Studies such as \cite{jie2019, sarzynska2021} exploit pretrained contextual embeddings from these models to achieve strong results in sequence labeling. Xu \cite{xu2021} further demonstrated that combining BiLSTM with BERT contextual embeddings yields notable improvements on OntoNotes 5.0. Labrak and Dufour \cite{Labrak2022} achieved unprecedented performance on POS tagging by leveraging Flair embeddings \cite{Akbik2019} integrated with BiLSTM.

\textbf{BiLSTM and its variants for sequence labeling.}
Bidirectional LSTMs (BiLSTMs) have long been a powerful architecture for modeling sequential dependencies among words in sequence labeling tasks such as NER \cite{Ghaddar2018, ma2016, Plank2016}. However, BiLSTM representations lack sufficient global context for inner tokens. To address this, Liu \cite{Liu2019} proposed the GCDT architecture that deepens the recurrent state transitions and assigns a global state per token. Meng and Zhang \cite{Meng2019} enhanced hidden-to-hidden transitions via multiple nonlinear transformations, achieving state-of-the-art results on WMT14. Zhang et al. \cite{Zhang2018} proposed S-LSTM, which assigns a shared global sentence representation at each time step while enabling parallel computation of token representations. Xu et al.\cite{XuBetterFeatureIntegration} introduced Synergized-LSTM (Syn-LSTM) to combine contextual features with structural information extracted via Graph Neural Networks (GNNs).

\textbf{Gate Mechanisms.}
Gate mechanisms are widely employed to control vector information flow. In standard LSTMs, gates regulate the influence of history and current inputs in cell computations. Chen et al. \cite{Chen2019} utilize gated relational neural networks to capture long-range dependencies, while Yuan et al. \cite{Yuan2020} and Zeng et al. \cite{Zeng2016} apply gate-based control to fuse multi-scale semantic information for object detection.

\begin{figure*}[]
\centering
\captionsetup{justification=centering}

\includegraphics [width= \textwidth, height=10cm] {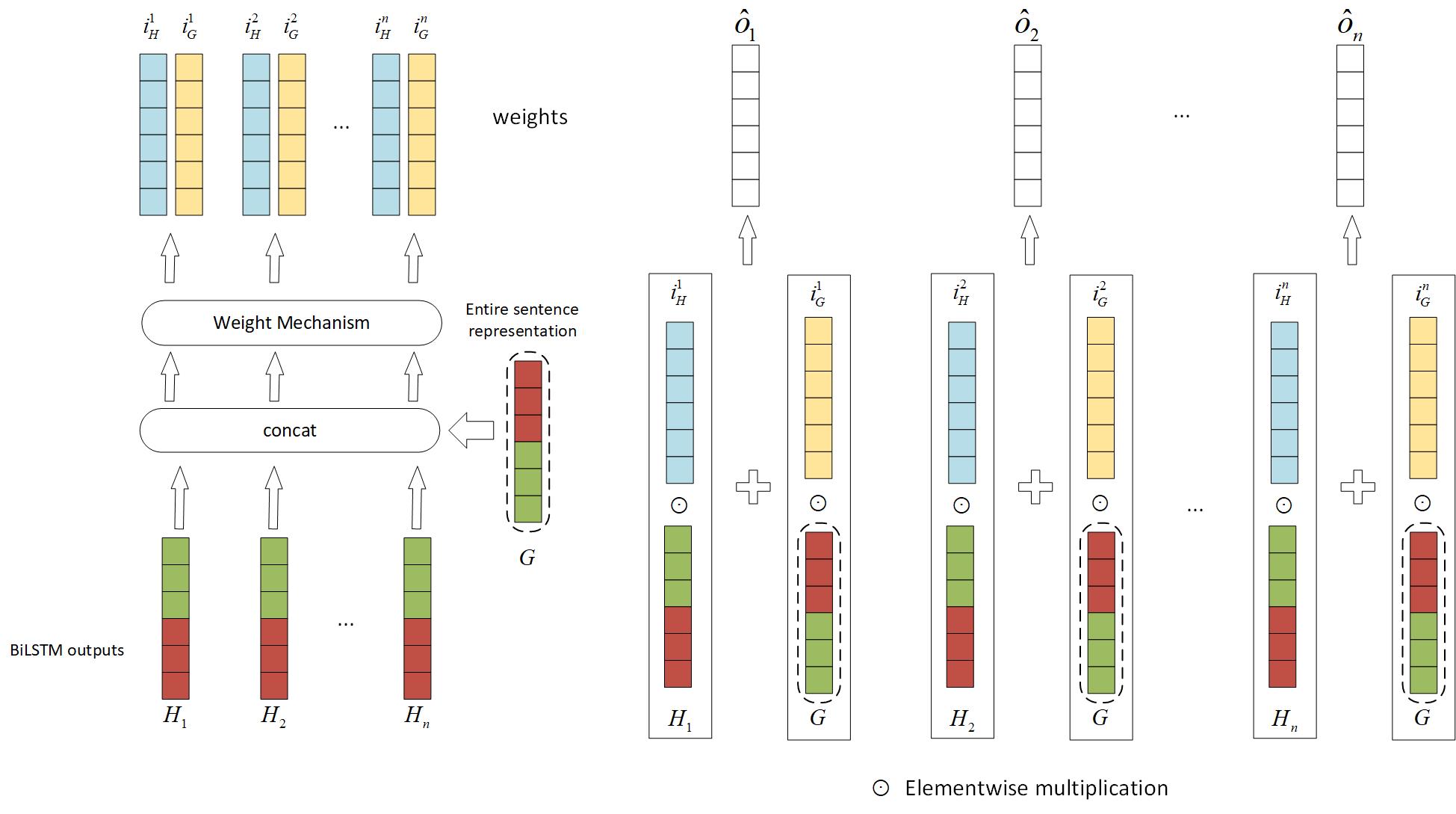}
\caption{Architecture details of The Global Context Mechanism}
\label{fig_2}
\end{figure*}

\section{Model}
The baseline neural architecture employed in this work, illustrated in Figure \ref{fig_1}, consists of two main components: a pretrained transformer for generating contextualized word representations, and a BiLSTM to capture sequential dependencies. Our proposed global context mechanism can be applied either after the BiLSTM layer or directly on the output of the pretrained transformer to augment the representations with global sentence-level information.

In the following subsections, we first describe the process of obtaining contextualized word embeddings from pretrained transformers and modeling sequential information via BiLSTM. Then, we detail our global context mechanism. Finally, we briefly review the self-attention mechanism \cite{Vaswani2017} and Conditional Random Fields (CRF) as applied to sequence labeling.

\subsection{Self-Attention}

Self-attention architecture \cite{Vaswani2017} was proposed to achieve parallelization and reduce training time for long sequences with large vector dimensions. Compared to recurrent neural networks (RNNs) such as LSTMs or GRUs, self-attention constructs word representations by weighing the importance of different words in a parallel manner. For sequence labeling, this mechanism can remedy the XOR problems caused by BiLSTM’s lack of sufficient global sentence information, enabling the model to capture complex relationships and dependencies between words \cite{Li2020}.

Self-attention consists of three main components: Linear Transformation Layers, Scaled Dot-Product Attention, and Multi-Head Attention.

\textbf{Linear Transformation Layers}: These layers generate the query, key, and value vectors from the input representations.

\textbf{Scaled Dot-Product Attention}: This module computes attention weights by calculating the dot product between queries and keys, scales it by the square root of the key dimension, normalizes via softmax, and uses these weights to aggregate value vectors, thus generating contextualized representations.

\textbf{Multi-Head Attention}: Multiple parallel attention heads compute different attention distributions; their outputs are concatenated and linearly transformed to form the final representation.

Formally, given a sentence representations $ H={H_1, H_2, ..., H_n} $ yield by BiLSTM, the linear transformation layer maps each $H_i \in H$ to the $query_i, key_i, value_i$ vectors:

\begin{equation}
Q=HW_Q, K=HW_K, V=HW_V  
\end{equation}

The Scaled Dot-Product Attention is calcuated as:
\begin{equation}
Attention(Q,K,V)=softmax(\frac{QK^T}{\sqrt{d_k}})V
\end{equation}
where $d_k$ is the dimension of the key vectors.
Finally, the outputs of each of the h attention heads are concatenated and projected by a weight matrix $ W_o $. 
\begin{equation}
MultiHead(Q,K,V)=Concat(head_1, head_2, ..., head_h)Wo
\end{equation}

\subsection{Pretrained Transformers}

Pretrained transformers are widely used as semantic encoding modules to generate deep contextualized representation for each word. Compared to traditional statistic word representations such as word2vector\cite{Mikolov2013}, Glove\cite{Pennington2014} and one-hot encodings, pretrained transformers produce representation that are more informative and dynamically adaptive to the specific context in which a word appears. 

Pretrained transformers are typically composed of multiple identical encoder blocks stacked on top of each other. Each encoder block contains four main components 1). self-attention mechanism 2). Add and Norm 3). Feed Forward Operation 4). Add and Norm

Self-Attention
Given a sequence of words $S=\{w_1,w_2,…,w_n\}$, where n denotes the length of the input sentence, Self-attention enables the model to attend to different positions of $S$ for a specific word $w_i$ and generated $H=\{h_1,h_2,...h_n\}$ with rich contextual information.
\begin{equation}
    H = self-attention(S)
\end{equation}
Add \& Norm (First)
After computing the output of the self-attention layer, it is added to the original input using a residual connection. This helps with gradient flow in deep networks. Layer normalization is then applied. 
\begin{equation}
 \mathbf{Z}_1 = \text{LayerNorm}(\text{Attention}(Q, K, V) + \mathbf{S})   
\end{equation}

where the Layer Normalization is defined as:
\begin{equation}
   \text{LayerNorm}(\mathbf{z}) = \gamma \cdot \frac{\mathbf{z} - \mu}{\sqrt{\sigma^2 + \epsilon}} + \beta
\end{equation}
with learnable parameters $ \gamma $ and $ \beta $, and $ \mu $, $ \sigma^2 $ being the mean and variance across the feature dimensions.

Feed Forward Network(FFN)
This is a position-wise fully connected feed-forward network, which consists two linear operation with a ReLU activation in between. 
\begin{equation}
    \text{FFN}(\mathbf{Z}_1) = W_2 \cdot \text{ReLU}(W_1 \mathbf{Z}_1 + b_1) + b_2
\end{equation}

where $ W_1 \in \mathbb{R}^{d_{ff} \times d}, W_2 \in \mathbb{R}^{d \times d_{ff}} $, and $ d_{ff} $ is the hidden layer dimension (usually larger than $ d $).

Add \& Norm (second)

Similar to the first Add \& Norm step, the output of FNN is added to the its input $Z_1$, and then followed by another layer normaliztion. 
\begin{equation}
    \mathbf{Z}_2 = \text{LayerNorm}(\text{FFN}(\mathbf{Z}_1) + \mathbf{Z}_1)
\end{equation}

At last, stacking this encoder blocks to get deep contextual representation for each word
\begin{equation}
    \mathbf{H^t} = EncoderBlock(\mathbf{H^{t-1}}), \quad for t=1,2,...,n
\end{equation}

\begin{figure*}[t]
\centering
\captionsetup{justification=centering}

\includegraphics [width= \textwidth] {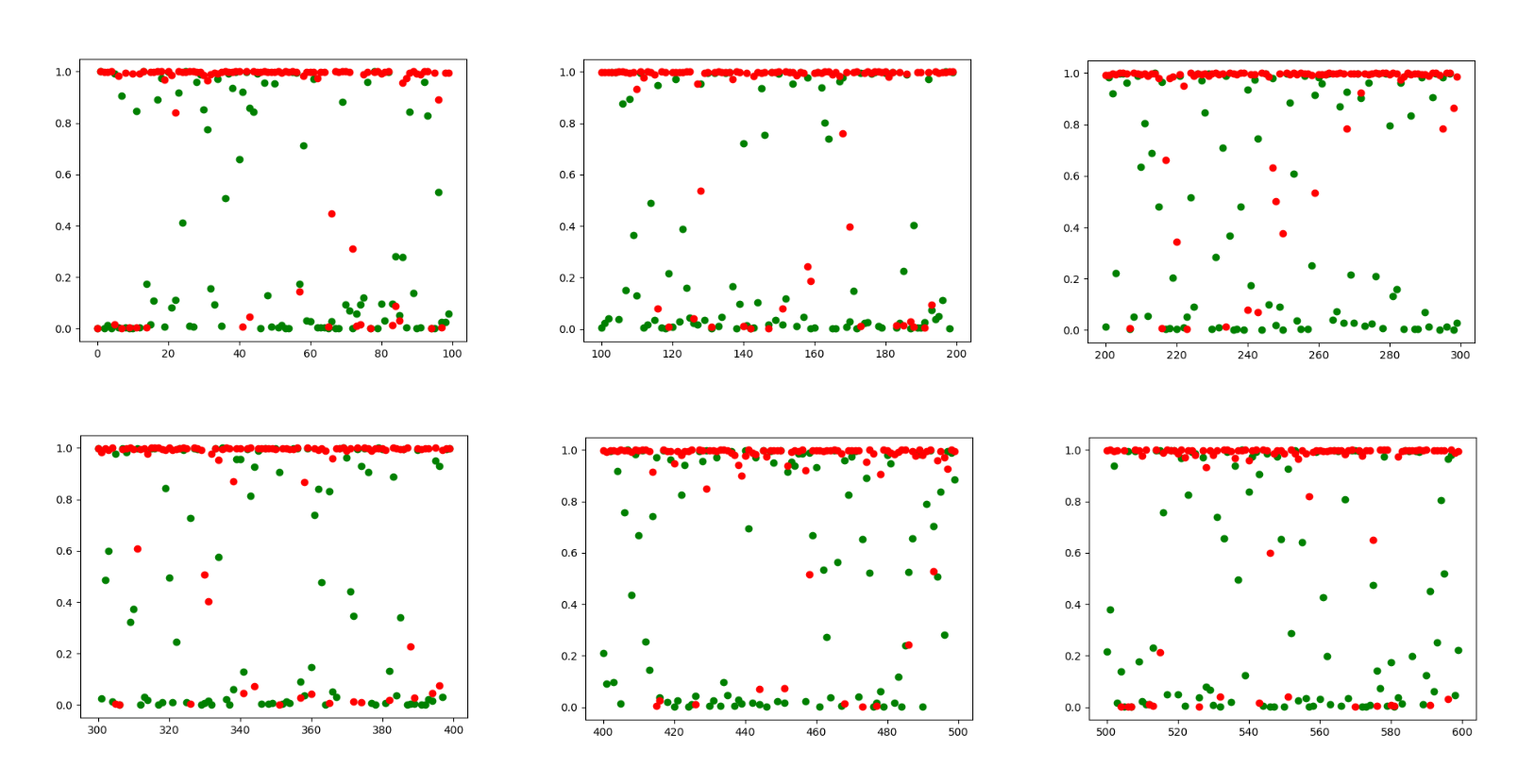}
\caption{Visualization for weights. Weights for context and BiLSTM are in green and red respectively.}
\label{fig_3}
\end{figure*} 

\subsection{BiLSTM} 
Bi-directional Long Short-Term Memory (BiLSTM) is employed to capture and enhance the sequential information within a sentence. In each BiLSTM cell, the forward hidden state and backward hidden state at the same time step are concatenated to form the final representation. This design helps incorporate information from both past and future contexts around each token.

However, the BiLSTM's ability to represent "global sentence information" remains limited. While the forward state at the last time step and the backward state at the first time step contain some global context, the intermediate time steps rely mainly on local sequential dependencies and do not fully model the global sentence-level information.

For instances, at time step t, The BiLSTM's cell generates the sentence representation $H_t$ based on the input sequence $Z2=\{z_1,z_2,…,z_n\}$:

\begin{equation}
\overrightarrow{H_t} = \overrightarrow{LSTM_t}(\overrightarrow{H_{t-1}},z_t)
\end{equation}

\begin{equation}
\overleftarrow{H_t} = \overleftarrow{LSTM_t}(\overleftarrow{H_{t+1}},z_t)
\end{equation}

\begin{equation}
H_t = \overrightarrow{H_t}\parallel\overleftarrow{H_t}
\end{equation}

Using word representations with only "weak global sentence information" can lead to XOR problems in sequence labeling tasks \cite{Li2020}. To illustrate, consider the following four phrases: (1) "Key and Peele" (work-of-art), (2) "You and I" (work-of-art), (3) "Key and I", and (4) "You and Peele". The first two are valid work-of-art entities, while the latter two are not. BiLSTM models can correctly label at most three out of these four examples. This limitation arises because BiLSTM lacks sufficient global context to disambiguate such cases, and it cannot be effectively overcome by simply increasing model parameters or stacking additional BiLSTM layers.

\subsection{Global Context Mechanism}

The global context mechanism fuse the global sentence information with local representation of each word without diminishing the original local information. This is achieved by applying only two additional layers on top of both BiLSTM and pretrained transformers. 

As show in Figure\ref{fig_2}, the global context mechanism first generate attention weights for global context representation and local context representation and then performs an element-wise fusing of these two representations. 

Specifically, for BiLSTM, the forward representation at the last time step and the backward representation at the first time step are taken as the forward global sentence representation and backward global sentence representation, respectively. These representations aggregate information from the entire sentence. For pretrained transformers, the [CLS] and [SEP] tokens are used to represent the forward global and backward global sentence representations.

In our model, the computation process is as follows:

 Given the sentence representation $H=\{H_i,H_2,…,H_n\} $ where $H \ni R^{n×d}$ generated either by BiLSTM or pretrained transformers. we denote the forward  global sentence representation and backward global representation as $ G=\overrightarrow{H_n}\parallel\overleftarrow{H_1} $. 
 
For the representation at $ t_{th} $ position, we concatenat its local representation and the global sentence representation to derive $K^t$ as the key for gate-weight mechanism firstly.

\begin{equation}
  \widehat{K^t}  = G \parallel H_t
\end{equation}

And then we conduct a feed forward operation to extract relevant features from $K_t$ for local representation at $ t_{th} $ and global sentence representation.

\begin{equation}
R_H = W_HK_t + b_H
\end{equation}
\begin{equation}
R_G = W_Gk_t + b_G
\end{equation}

Here, $W_H$ and $W_G \in R^{2d×d}$; $R_G^t$, $R_G^t$ correspond to global information $G$ and the current sentence representation $H_i$ respectively. 

Base on the $R_H$, $R_G$, we generated element-wise weights $i_H^t$ and $i_G^t$ by a sigmoid function for the $t_{th}$ local word representation and global sentence representation.
\begin{equation}
i_H^t = sigmoid(R_H^t)
\end{equation}
\begin{equation}
i_G^t = sigmoid(R_G^t)
\end{equation}
At the final step, it fuses the local word representation and global sentence representation according supplementary features of BiLSTM to get the final representation. 

\begin{equation}
\widehat{H_t} = i_H^t \odot H_t \parallel i_G^t \odot G
\end{equation}
Here, $\odot$ denotes element wise product.

\subsection{Conditional Random Fields (CRFs)}

Conditional Random Fields (CRFs) \cite{Lafferty2001} are a class of discriminative probabilistic models commonly used for structured prediction tasks, especially in natural language processing and sequence labeling problems. Unlike generative models such as Hidden Markov Models (HMMs), CRFs directly model the conditional probability distribution $ P(Y \mid X) $, where $ Y $ is the output label sequence and $ X $ is the corresponding input sequence.

In a linear-chain CRF, which is typically applied to sequential data, the dependencies between labels are captured through transition features, while the relationship between inputs and outputs is modeled using state features. The conditional probability of a label sequence given an input sequence is defined as:

\begin{equation}
    P(Y \mid X; \theta) = \frac{1}{Z(X)} \exp\left( \sum_{t=1}^{T} \sum_{k=1}^{K} \theta_k f_k(y_t, y_{t-1}, x_t) \right)
\end{equation}

where $ \theta_k $ are the model parameters, $ f_k(\cdot) $ are feature functions, $ T $ is the sequence length, and $ Z(X) $ is the normalization factor (partition function) that ensures the probabilities sum to one.

Training a CRF involves maximizing the log-likelihood of the training data with respect to the model parameters $ \theta $, often using optimization techniques such as L-BFGS or stochastic gradient descent. During inference, the Viterbi algorithm is typically employed to find the most probable label sequence given an input sequence.

\subsection{Classification}
The last module of the model is classification module. It computes the corresponding label for each word as follows:

Give the representations $ {\widehat{H_1}, \widehat{H_2}, ..., \widehat{H_n}} $ for all time steps, we compute prediction outcomes using a softmax operation. results by adopting soft-max operation.

\begin{equation}
\tilde{O_t} = softmax(W_c \widehat{H_t})
\end{equation}

\section{Experiments}

\subsection{ Dataset and Metric}
\begin{table}[h]
\centering
\caption{Statistics of ABSA dataset}
\label{tab:absa_info}
\begin{tabular}{llrrrr}
\hline
                         & Dataset   & Train & Dev & Test & Total \\ \hline
\multirow{2}{*}{LAPTOP}  & \# sent   & 2741  & 304 & 800  & 4245  \\
                         & \# aspect & 2041  & 256 & 634  & 2931  \\
                         \hline
\multirow{2}{*}{REST 14} & \# sent   & 2737  & 304 & 800  & 3841  \\
                         & \# aspect & 3205  & 329 & 1102 & 4636  \\
                         \hline
\multirow{2}{*}{REST 15} & \# sent   & 1183  & 130 & 686  & 1999  \\
                         & \# aspect & 1069  & 127 & 539  & 1735  \\
                         \hline
\multirow{2}{*}{REST 16} & \# sent   & 1800  & 200 & 676  & 2676  \\
                         & \# aspect & 1569  & 166 & 594  & 2329  \\ \hline
\end{tabular}
\end{table}

\textbf{ABSA}. E2E-ABSA is tagged with three aspect sentiment types (POS, NEG, NEU). In this work, we use four datasets originating from SemEval \cite{pontiki2014,pontiki-etal-2015-semeval,pontiki2016} but re-prepared by Li et al. \cite{li2019b}. The statistics of these four dataset are summarized in Table\ref{tab:absa_info}. 

\textbf{NER}. we use two Enlgish dataset CoNLL2003, Wnut2017 and a Chinese dataset Weibo in our experiments. CoNLL2003 and Wnut2017 are tagged with four linguistic entity types (PER, LOC, ORG, MISC) and six linguistic entity types ('product', 'group', 'person', 'corporation', 'work', 'location'). Weibo are tagged with ('ORG.NOM', 'PER.NOM', 'LOC.NOM', 'GPE.NAM', 'LOC.NAM', 'GPE.NOM', 'ORG.NAM', 'PER.NAM'). The summary of these dataset are shown in Table\ref{tab:ner_info}.

\begin{table}[h]
\centering
\caption{Statistics of NER datasets}
\label{tab:ner_info}
\begin{tabular}{llrrrr}
\hline
                           & Dataset   & Train & Dev  & Test & Total \\
\hline
\multirow{2}{*}{CoNLL2003} & \# sent   & 14987 & 3466 & 3684 & 22137 \\
                           & \# entity & 23222 & 5861 & 5584 & 34677 \\
\hline
\multirow{2}{*}{Wnut2017}  & \# sent   & 3394  & 1009 & 1287 & 5690  \\
                           & \# entity & 1937  & 819  & 1050 & 3806  \\
\hline
\multirow{2}{*}{Weibo}     & \# sent   & 1350  & 270  & 270  & 1890  \\
                           & \# entity & 1633  & 325  & 358  & 2316 \\
\hline
\end{tabular}
\end{table}

\textbf{Metric} We use the traditional BIO tagging system and choose “relaxed” micro averaged F1-score, which regards a prediction as the correct one as long as a part of NE is correctly identified. This evaluation metric has been used in several related publications, journals, and papers on NER [neural architectures for named entity recognition] [Bidirectional LSTM-CRF Models for Sequence Tagging] [Use of support vector machines in extended named entity recognition] [End-to-end sequence labeling via bi-directional LSTM-CNNs-CRF].

\subsection{Experiments Setting}
The overall model architecture is illuatrated in Figure\ref{fig_1}, where the global context mechanism add either after pretrained transformers and BiLSTM. In this work we employ Bert-Base-cased and Roberta(RoBERTa: A Robustly Optimized BERT Pretraining Approach) as base pretrained transformers for all ABSA-E2E datasets and Conll2003. For the WNUT2017 dataset, we use Bert-twitter(BERTweet: A pre-trained language model for English Tweets) and Roberta-base. For the Weibo datset, we utilize Bert-base-Chinese and MacBert(Revisiting Pre-Trained Models for Chinese Natural Language Processing).

Training was performed on an NVIDIA Tesla V100 GPU, using with 32GB memory, with the Adam optimizer. An early Early stopping strategy based on validation performance was applied to prevent overfitting. To mitigate the of learning rate, batch size and dropout rate, we trained models across a range of learning rates for each experiment and selected the model with best F1 score as model final performance metric. Detailed information on the learning rate sets is provided in the appendix.

\subsection{Main Results}



\begin{table*}
  \centering
\begin{tblr}{
  width = \linewidth,
  colspec = {Q[120]Q[80,c]Q[80,c]Q[80,c]Q[80,c]Q[80,c]Q[80,c]Q[80,c]},
  column{even} = {r},
  column{3} = {r},
  column{5} = {r},
  column{7} = {r},
  hline{1,10} = {-}{0.08em},
  hline{2,6} = {-}{0.05em},
}
Models         & Laptop14           & Restaurant14    & Restaurant15    & Restaurant16       & Conll2003       & Wnut2017        & WeiboNER           \\
BERT           & 58.32              & 71.75           & 57.59           & 66.93              & 91.49           & 53.77           & 69.69              \\
+ context      & 60.35$\uparrow$    & 72.45$\uparrow$ & 60.31$\uparrow$ & 70.98$\uparrow$    & 91.72$\uparrow$ & 54.78$\uparrow$ & 69.00$\downarrow$ \\
BERT-BiLSTM    & 61.33              & 74.00           & 61.58           & 71.47              & 91.76           & 55.76           & 69.53              \\
+ context      & 61.85$\uparrow$     & 75.31$\uparrow$ & 61.65$\uparrow$ & 70.76$\downarrow$ & 91.84$\uparrow$ & 56.26$\uparrow$ & 72.08$\uparrow$    \\
RoBERTa        & 68.43              & 77.38           & 66.97           & 75.47              & 92.46           & 58.29           & 69.88              \\
+ context      & 69.27$\uparrow$     & 78.17$\uparrow$ & 68.34$\uparrow$ & 76.20$\uparrow$    & 92.61$\uparrow$ & 59.27$\uparrow$ & 71.17$\uparrow$    \\
RoBERTa-BiLSTM & 69.52              & 77.52           & 68.54           & 76.90              & 92.52           & 59.58           & 70.28              \\
+ context      & 65.15$\downarrow$ & 78.29$\uparrow$ & 69.37$\uparrow$ & 77.30$\uparrow$     & 92.71$\uparrow$ & 59.61$\uparrow$ & 70.30$\uparrow$    
\end{tblr}
\caption{Performance Comparison with and without Global Context on ABSA and NER Datasets}
\label{tab:f1_all}
\end{table*}

The overall results of adding global context mechanism after BiLSTM and pretrained transformers are presented in Table\ref{tab:f1_all}. After incorporating the global context mechanism, the F1 score of BERT and RoBERTa improved across all datasets, except for BERT on WeibNER. Notably, BERT exhibited significant F1 improvements on Restaurant16, Restaurant15, Laptop14 and WNUT2017, with gains of $ 4.05 $, $ 2.52 $, $2.03$ and $1.01$, respectively. For RoBERTa, the F1 improvements on Restaurant15, WeiboNER,  and WNUT2017 were also competitive,  with  $1.34$, $1.29$ and $0.98$. 

\begin{table}[htpb]
    \centering
\begin{tblr}[]{
  width = \linewidth,
  colspec = {Q[294,c]Q[306,c]Q[319,c]},
  hline{1,6} = {-}{0.08em},
  hline{2} = {-}{},
}
Model          & Train Speed(it/s) & Inference Time(s) \\
+context       & 4.47              & 1.95              \\
+CRF           & 3.7               & 2.71              \\
BiLSTM+context & 4.19              & 2.01              \\
BiLSTM+CRF     & 3.57              & 2.82              
\end{tblr}
    \caption{Efficiency comparasion of Global Context Mechanism and CRF}
    \label{tab:crf_global}
\end{table}

When the global context mechanism was applied after BiLSTM, F1 improvements observed on all datasets except for BERT on Restaurant 16 and Roberta on BiLSTM. In these cases, better F1 scores achieved by adjusting  the position of forward and backward global sentence representation. In particular, for BERT-BiLSTM model, WeiboNER and Restaurant16 shoed competitive improvements improvements $2.55$ and $1.31$, with third highest score on WeiboNER benchmark $72.08$. For Roberta-BiLSTM, Restaurant15 and Restaurant14 achieved competitive gains of $0.83$ and $0.77$.

We report training speed (iteration per seconds), inference time and parameter cost of global context mechanism on the Restaurant14 dataset. Specifically, we average the training speed and inference time over validation set over three runs to obtain the final metric. As shown in table\ref{tab:per_statistics}, the global context mechanism increase the parameters count by approximately $2\%$and $1.3\%$ when combined with BERT, RoBERTa, BERT-BiLSTM and RoBERTa-BiLSTM, while incurring minimal training cost. Regrading inference, it adds a a maximum overhead of 0.16 seconds. These statistics demonstrate the efficiency of the global context mechanism in terms of memory usage, training speed and inference time.

\begin{table}[htbp]
    \centering
\begin{tblr}[
  caption = {Impact of Global Context Mechanism on Training Efficiency, Inference Speed, and Model Complexity\label{tab:per_statistics}},
  label = {tab:table5},
]{
  width = \linewidth,
  colspec = {Q[213]Q[110]Q[110]Q[110]},
  hline{1-2,6,10} = {-}{},
}
Models        & Num Params(M) & Train Speed(it/s) & Inference Time(s) \\
BERT          & 108.3         & 4.52              & 2.0               \\
+ context     & +2.5          & -0.07             & -0.05             \\
BERT-BiLSTM   & 110.8         & 4.41              & 1.95              \\
+context      & +1.5          & -0.1              & +0.06             \\
RoBERTa       & 124.6         & 4.61              & 2.0               \\
+context      & +2.4          & -0.12             & -0.07             \\
RoBERTa-BiLSTM & 127.2         & 4.45              & 1.92              \\
+context      & +1.5          & -0.12             & +0.16             
\end{tblr}
    \caption{Impact of Global Context Mechanism on Training Efficiency, Inference Speed, and Model Complexity}
    \label{tab:per_statistics}
\end{table}

\section{Analysis}

\begin{table*}[htbp]
\centering
\begin{tblr}{
  width = \linewidth,
  colspec = {Q[80,l] Q[88,l] Q[120,c] Q[120,c] Q[80,c] Q[80,c] Q[65,c] Q[120,c] Q[120,c]},
  row{1} = {c, font=\bfseries},
  cell{1}{1} = {c=2}{0.146\linewidth},
  cell{2}{1} = {r=4}{},
  cell{2}{3} = {c},
  cell{2}{4} = {c},
  cell{2}{5} = {c},
  cell{2}{6} = {c},
  cell{2}{7} = {c},
  cell{2}{8} = {c},
  cell{2}{9} = {c},
  cell{3}{3} = {c},
  cell{3}{4} = {c},
  cell{3}{5} = {c},
  cell{3}{6} = {c},
  cell{3}{7} = {c},
  cell{3}{8} = {c},
  cell{3}{9} = {c},
  cell{4}{3} = {c},
  cell{4}{4} = {c},
  cell{4}{5} = {c},
  cell{4}{6} = {c},
  cell{4}{7} = {c},
  cell{4}{8} = {c},
  cell{4}{9} = {c},
  cell{5}{3} = {c},
  cell{5}{4} = {c},
  cell{5}{5} = {c},
  cell{5}{6} = {c},
  cell{5}{7} = {c},
  cell{5}{8} = {c},
  cell{5}{9} = {c},
  cell{6}{1} = {r=4}{},
  cell{6}{3} = {c},
  cell{6}{4} = {c},
  cell{6}{5} = {c},
  cell{6}{6} = {c},
  cell{6}{7} = {c},
  cell{6}{8} = {c},
  cell{6}{9} = {c},
  cell{7}{3} = {c},
  cell{7}{4} = {c},
  cell{7}{5} = {c},
  cell{7}{6} = {c},
  cell{7}{7} = {c},
  cell{7}{8} = {c},
  cell{7}{9} = {c},
  cell{8}{3} = {c},
  cell{8}{4} = {c},
  cell{8}{5} = {c},
  cell{8}{6} = {c},
  cell{8}{7} = {c},
  cell{8}{8} = {c},
  cell{8}{9} = {c},
  cell{9}{3} = {c},
  cell{9}{4} = {c},
  cell{9}{5} = {c},
  cell{9}{6} = {c},
  cell{9}{7} = {c},
  cell{9}{8} = {c},
  cell{9}{9} = {c},
  vline{2,7} = {2-6,7-9}{},
  vline{7} = {3-5,7-9}{},
  hline{1-2,6,10} = {-}{},
  stretch = 1.2,
}
Models  &                & Laptop14          & Restaurant14      & Restaurant15 & Restaurant16 & Conll2003 & Wnut2017          & WeiboNER          \\
BERT    & \hbox{self-attention} & 61.85             & 72.33             & 59.56        & 70.39        & 91.66     & 55.08             & 67.72             \\
        & wo [C][S]      & 61.11$\downarrow$ & 72.58             & 59.66        & 69.84        & 91.6      & 54.33$\downarrow$ & 67.85             \\
        & context        & 60.35             & 72.45             & 60.31        & 70.98        & 91.72     & 54.78             & 69.00             \\
        & wo [C][S]      & 59.48             & 72.7              & 61.35        & 70.84        & 91.75     & 54.93             & 68.97             \\
RoBERTa & \hbox{self-attention} & 69.92             & 78.45             & 68.15        & 76.67        & 92.58     & 57.79             & 68.9              \\
        & wo [C][S]      & 69.36$\downarrow$ & 77.05$\downarrow$ & 68.31        & 76.42        & 92.83     & 57.96             & 67.85$\downarrow$ \\
        & context        & 69.27             & 78.17             & 68.34        & 76.20        & 92.61     & 59.27             & 71.17             \\
        & wo [C][S]      & 67.69$\downarrow$ & 77.94             & 67.96        & 76.72        & 92.64     & 58.96             & 70.76             
\end{tblr}
\caption{Impact of special tokens, [C] and [S] denote [CLS] and [SEP].}
\label{tab:f1_sepcial}
\end{table*}
In this section, we conduct an in-depth analysis of: 1) the comparison between the global context mechanism and two classical modules for sequence labeling—self-attention and CRF; 2) the impact of using [CLS] and [SEP] as global sentence information; 3) the methods of concatenating local and global representations; and 4) the comparison between the global context mechanism and stacked BiLSTM layers. 

\subsection{Comparison with CRF and Self-Attention}
1. Comparison with Self-Attention. When combined with BERT and RoBERTa on the ABSA-E2E task, both self-attention and the global context mechanism improve the F1 score compared to the original pretrained transformer on most datasets except Weibo. For the ABSA-E2E task, the performance of self-attention and the global context mechanism is generally competitive. Specifically, when using with BERT, the global context mechanism outperforms self-attention on an all datasets except except Laptop14. Conversely, when used with RoBERTa, self-attention achieves higher F1 scores on all ABSA-E2E datasets except Retaurant15. However, for NER tasks, the global context mechanism clearly outperforms self-attention. self-attention only surpasses the global context mechanism on Wnut2017 with BERT, and it achieves lower F1 score than the original baseline on Weibo NER (for both BERT and RoBERTa) and Wnut2017 with RoBERTa. we attribute this difference to the nature of the tasks: in ABSA-E2E, each entity's tag is combined with a sentiment label such as 'NER', 'POS', 'NEU', which is strongly associated with global sentence information. In contrast, NER entity tags depended more on local contextual information. This suggests that, compared to the more aggressive attention mechanism that fuse word-level representations, the global context mechanism effectively incorporates global sentence information while preserving local details. When used with BiLSTM, there is a noticeable F1 score difference between self-attention and the global context mechanism, especially in NER tasks. This indicates that, compared to traditional attention mechanism, the global context mechanism better retains the sequential information generated by BiLSTM.

2. Comparison with CRF. Compared to CRF, integrating the global context mechanism achieves competitive F1 scores on most datasets. When added after BiLSTM on NER tasks, the global context mechanism generally shows greater F1 improvement than CRF, except for CoNLL2003. For ABSA-E2E, the global context mechanism achieves better F1 scroes than CRF when combined with RoBERTa, except on Laptop14. With BERT, CRF yields slightly better F1 scores than the global context mechanism, though the gap  is very small. Regarding efficiency, as shown in Table\ref{tab:crf_global}( measured on BERT on Restaurant14 dataset), the training speed of global context mechanism is faster than CRF by $20\%$ and inference time up to $28\%$ faster.

\subsection{Impact of Special Tokens}

We evaluated the effect of [CLS] and [SEP] special tokens in both self-attention and the global context mechanism across on all seven datasets. As shown in Table\ref{tab:f1_sepcial}, we observed a decrease in F1 score in 16 out of 28 experiments when these two special toknes were not used. Specifically, substantial F1 degradation—greater than $ 0.5 $—occurred for self-attention combined RoBERTa on Laptop14, Restaurant14 and Weibo NER, as well with BERT on Laptop14 and Wnut2017. For the global context mechanism, a noticeable F1 drop appeared on Laptop14 with RoBERTa.

We attribute this sensitivity to the special tokens in ABSA-E2E tasks, where entity tags such as 'NEG', 'NEU', 'POS' rely heavily on global sentence-levle information. The global sentence-level information encoded in the [CLS] and [SEP], which is benefical for sentence classification tasks, also aids the ABSA-E2E tasks. 

Furthermore, we found that self-attention is more sensitive to the presence of special tokens compared to the global context mechanism. There are five of the six significant F1 drops comes from using with self-attention.

\subsection{Comparison with adding parameters}
We also compared the global context mechanism with the approach of simply adding parameters. For pretrained transformers, we added extra fully connected layers, and for BiLSTM, we increased the number of stacked layers. The results, show in Table\ref{tab:f1_comparison}, indicate that adding parameters caused F1 drops in 12 out of 14 experiments for pretrained transformers except on Conll2003 and Restaurant15 combined with BERT. Notably, sinigificant F1 drops occurred across all ABSA-E2E datasets combined with RoBERTa, as well as in Weibo NER combined with BERT. 

For BiLSTM, 12 out of 14 experiments showed F1 decreases, except for Laptop14 combined with BERT and Restaurant15 combined RoBERTa. We also observed large F1 dorps on Restaurant16, WNUT2017 and Weibo NER with RoBERTa, and on Restaurant15, Restaurant16 and Wnut2017 with BERT.

These results suggest that the lack of global sentence-level information cannot be remedied simply by adding parameters. Moreover, increasing parameters may negatively affect the original structural information present in the original representations.

\begin{table*}[htbp]
    \centering

\begin{tblr}{
  width = \linewidth,
  colspec = {Q[152]Q[96]Q[125]Q[125]Q[125]Q[100]Q[100]Q[104]},
  column{even} = {c},
  column{3} = {c},
  column{5} = {c},
  column{7} = {c},
  hline{1-2,12,22} = {-}{},
  hline{7,17} = {-}{dotted,red},
}
Models           & Laptop14       & Restaurant14   & Restaurant15   & Restaurant16   & Conll2003      & Wnut2017       & WeiboNER       \\
BERT             & 58.32          & 71.75          & 57.59          & 66.93          & 91.49          & 53.77          & \textbf{69.69} \\
+ linear         & 58.12          & 70.19          & 59.13          & 66.46          & 91.84          & 53.41          & 64.35          \\
+ context        & 60.35          & 72.45          & 60.31          & 70.98          & 91.72          & 54.78          & 69.00          \\
+ CRF            & 60.3           & \textbf{72.76} & \textbf{60.76} & \textbf{71.47} & \textbf{91.85} & \textbf{55.16} & 68.72          \\
+ self-attention & \textbf{61.85} & 72.33          & 59.56          & 70.39          & 91.66          & 55.08          & 67.72          \\
BERT-BiLSTM      & 61.33          & 74.00          & 61.58          & 71.47          & 91.76          & 55.76          & 69.53          \\
+ context        & 61.85          & \textbf{75.31} & 61.65          & 70.76          & 91.84          & \textbf{56.26} & \textbf{72.08} \\
+CRF             & 62.19          & 73.1           & \textbf{61.74} & 71.41          & \textbf{91.98} & 55.82          & 70.86          \\
+self-attention  & 60.77          & 73.35          & 61.72          & \textbf{71.59} & 91.58          & 54.56          & 67.81          \\
BiLSTM(2)        & \textbf{62.27} & 73.09          & 60.44          & 67.9           & 91.74          & 54.17          & 69.53          \\
Roberta          & 68.43          & 77.38          & 66.97          & 75.47          & 92.46          & 58.29          & 69.88          \\
+linear          & 67.31          & 76.33          & 62.05          & 72.43          & 92.41          & 57.97          & 69.01          \\
+ context        & 69.27          & 78.17          & 68.34          & 76.20          & 92.61          & 59.27          & \textbf{71.17} \\
+CRF             & \textbf{70.78} & 77.16          & \textbf{68.39} & \textbf{76.9}  & \textbf{92.62} & \textbf{59.31} & 70.15          \\
+self-attention  & 69.92          & \textbf{78.45} & 68.15          & 76.67          & 92.58          & 57.79          & 68.9           \\
Roberta-BiLSTM   & 69.52          & 77.52          & 68.54          & 76.90          & 92.52          & 59.58          & 70.28          \\
+ context        & 69.15          & \textbf{78.29} & \textbf{69.37} & \textbf{77.30} & 92.71          & \textbf{59.61} & \textbf{70.30} \\
+CRF             & \textbf{70.96} & 77.89          & 69.11          & 76.9           & \textbf{92.91} & 57.46          & 68.11          \\
+self-attention  & 69.92          & 77.88          & 68.34          & 76.01          & 92.73          & 58.89          & 68.89          \\
BiLSTM(2)        & 69.43          & 76.94          & 69.11          & 74.36          & 92.52          & 58.16          & 69.01          
\end{tblr}
    \caption{F1 score comparison between different architectures}
    \label{tab:f1_comparison}
\end{table*}

\subsection{Case studies}
Using Weibo NER as example, we present a case comparing the predicted tags from original BiLSTM and BiLSTM enhanced with the global context mechanism. As show in Table \ref{tab:table8}, the Chinese characters \begin{CJK*}{UTF8}{gbsn}‘大’ and ’师’ \end{CJK*}can either refer to an individual or signify a title, depending on the context. By leveraging the global context text mechanism, the model correctly assigns the relevant types for \begin{CJK*}{UTF8}{gbsn}‘大师’ \end{CJK*}, which was incorrectly predicted by the original BiLSTM.

We also visualize the weights of the local representation, denoted $i_H$ and the weights of the global sentence-level representation, denoted as $i_G$. The local and the global sentence representation are segregated into six segments at intervals of 100, followed by scatter plot visualization. As show in Figure\ref{fig_3}, the local representation weights are high at most positions, while only a few parts of the global vectors have larger weights.

This suggest that a small portion of the global sentence level information is sufficient to enhance local representation.

\section{Conclusion}

In this work, we proposed a global context mechanism to enhance to representation of individual words by incorporating global sentence information for both BiLSTM and pretrained Transformer. Compared with previously designed complex RNNs that time-consuming in implementation, training and inference, the global context mechanism offers a simpler and more efficient alternative. Additionally, it outperforms the classical CRF architecture, which are very useful to build sequential information between tags, in terms of training and inference efficiency. By integrating this mechanism with pretrained transformers, we achieve improvements of F1 score across seven sequence labeling datasets. Notably, we attained third-highest score on Weibo NER dataset without relying on any additional strategies.

\begin{CJK*}{UTF8}{gbsn}
\begin{table*}[]
\centering

\resizebox{0.6\textwidth}{4 cm}{%
\begin{tabular}{@{}llll@{}}
\toprule
Sentece & Gold Standard & context   & w/o context                       \\ \midrule
分       & O             & O         & O                                 \\
手       & O             & O         & O                                 \\
大       & O             & O         & \cellcolor[HTML]{FFFFC7}B-PER.NOM \\
师       & O             & O         & \cellcolor[HTML]{FFFFC7}I-PER.NOM \\
贵       & O             & O         & \cellcolor[HTML]{FFFFC7}B-PER.NOM \\
仔       & O             & O         & O                                 \\
邓       & B-PER.NAM     & B-PER.NAM & B-PER.NAM                         \\
超       & I-PER.NAM     & I-PER.NAM & I-PER.NAM                         \\
四       & O             & O         & O                                 \\
大       & O             & O         & O                                 \\
名       & O             & O         & O                                 \\
捕       & O             & O         & \cellcolor[HTML]{FFFFC7}I-PER.NOM \\ \bottomrule
\end{tabular}%
}
\caption{Weibo case analysis. The errors are in yellow.\label{tab:table8}}
\end{table*}
\end{CJK*}

\section{Limitations}

For other sequence labeling tasks, we perform simple experiments on two Part-of-Speech tagging tasks: Conll2003 and Universal Dependencies (UD) v2.11 (Silveira , 2014) using a heuristic learning rate, achieving only minor accuracy improvements. Considering the significant energy and time costs associated with tuning hyperparameters combinations, as well as the already competitive performance of BERT and RoBERTa on these datasets, we chose not to perform further experiments to reduce energy consumption and carbon emissions.

 Regarding hyperparameters, experiments were conducted across different GPU platforms, including Nvidia 1080Ti, A10, A100 32G and A100 16G. we found that the optimal hyperparameters vary depending on the training platform; therefore, the best learning rates reported here may not be reproducible on other hardware configurations. 

\bibliographystyle{IEEEtran}
\bibliography{export.bib}

\newpage

\section{Appendices}

\subsection{Learning Rate} 
To mitigate the impact of hyperparameters, we conduct experiments of each model over a range of parameters combinations, selecting the best F1 score as the final evaluation metric. As shown in Table, for BERT and RoBERTa we searched batch size [16, 30] and learning rates of {1e-5, 2e-5, 8-5}. When combining pretrained transformers with BiLSTM and the Context Mechanism, we simplified the learning rate search by reducing the transformer learning rate to {1e-5, 8e-5}. Similarly. for BiLSTM when used with the Context Mechanism, the learning rate set was narrowed from {5e-4, 1e-3, 5e-3} to {5e-4, 1e-3}, when using with context mechanism. For the global context mechanism, we also evaluated dropout rates of {0.1, 0.3}. The detailed parameters sets are listed below.

\begin{table}[htbp]
  \centering
  \small
  \begin{tabularx}{\linewidth}{c c c c c}
    \toprule
    Batch Size & BERT  & BiLSTM & Context Mechanism\textsuperscript{\footnotemark} & Dropout \\
    \midrule
    16         & 1e-5  & 5e-4   & 1e-4           & 0.1     \\
    30         & 2e-5  & 1e-3   & 5e-4           & 0.3     \\
               & 8e-5  & 5e-3   & 5e-3           &           \\
    \bottomrule
  \end{tabularx}
  \caption{Hyperparameters set in experiments}
  \label{tab:hyper}
\end{table}

\footnotetext{In this table, the 'Context Mechanism' refers to one of the following: self-attention, global context mechanism, or CRF (Conditional Random Fields).}

\end{document}